\pdfoutput=1
\documentclass[11pt]{article}

\usepackage[preprint]{coling}
\usepackage{times}
\usepackage{latexsym}

\usepackage[T1]{fontenc}
\usepackage[utf8]{inputenc}

\usepackage{microtype}

\usepackage{inconsolata}

\usepackage{graphicx}
\title{Phase Diagram of Vision Large Language Models Inference: A Perspective from Interaction across Image and Instruction}

\author{Houjing WEI\textsuperscript{1},Yuting SHI\textsuperscript{1}, Naoya Inoue\textsuperscript{1, 2}\\
1. Japan Advanced Institute of Science and Technology 2. RIKEN\\
\{houjing,s2210096\}@jaist.ac.jp, naoya.inoue.lab@gmail.com}

\begin{document}
\maketitle
\begin{abstract}
\textbf{V}ision \textbf{L}arge \textbf{L}anguage \textbf{M}odels (VLLMs) usually take input as a concatenation of image token embeddings and text token embeddings and conduct causal modeling.
However, their internal behaviors remain underexplored, raising the question of interaction among two types of tokens.
To investigate such multimodal interaction during model inference, in this paper, we measure the contextualization among the hidden state vectors of tokens from different modalities.
Our experiments uncover a four-phase inference dynamics of VLLMs against the depth of Transformer-based LMs, including (\textbf{I) Alignment}: In very early layers, contextualization emerges between modalities, suggesting a feature space alignment. (\textbf{II) Intra-modal Encoding}: In early layers, intra-modal contextualization is enhanced while inter-modal interaction is suppressed, suggesting a local encoding within modalities.
(\textbf{III) Inter-modal Encoding}: In later layers, contextualization across modalities is enhanced, suggesting a deeper fusion across modalities. (\textbf{IV) Output Preparation}: In very late layers, contextualization is reduced globally, and hidden states are aligned towards the unembedding space. 
% Our findings provide insight into accelerating the inference via reducing model parameters.

\end{abstract}

\section{Introduction}

Recently, instruction-tuned \textbf{L}anguage \textbf{M}odels (LMs) have demonstrated impressive performance on cross-modal tasks when incorporated with other modalities, mainly vision~\cite{10.5555/3666122.3668264,liu2024visual,liu2024improvedbaselinesvisualinstruction, zhu2024minigpt, chen2023minigptv2largelanguagemodel, merullo2022linearly}. These \textbf{V}ision \textbf{L}arge \textbf{L}anguage \textbf{M}odels (VLLMs) usually feed a concatenation of image tokens and text tokens into frozen Transformer-based LMs.~\citet{schwettmann2023multimodalneuronspretrainedtextonly} identify the multimodal neuron in Transformer MLPs and find that neurons from MLPs, not from outputs of the fine-tuned projector, translate image semantics into related text. Another following work in ~\citet{verma2024crossmodalprojection} verifies that domain-specific visual attributes are modeled by LMs while fine-tuning the projector of VLLMs. Although these works provide prominent findings about the inner workings of VLLMs, how the multimodal inputs interact against LMs layers is still unknown, leading to our main research question: How does interaction happen among multimodal inputs (image, text) in the LMs of VLLMs? 

% Existing work has shown that the transformation between visual representation and text embedding can be achieved via a learnable adapter layer~\citep{schwettmann2023multimodalneuronspretrainedtextonly, verma2024crossmodalprojection}, indicating implicit similarity between them.
% However, the inner workings of such semantically equivalent conversion between modalities are still unknown.

\begin{figure}
    \centering
    \begin{minipage}[b]{0.48\linewidth}
    \centering
    \includegraphics[width=\textwidth]{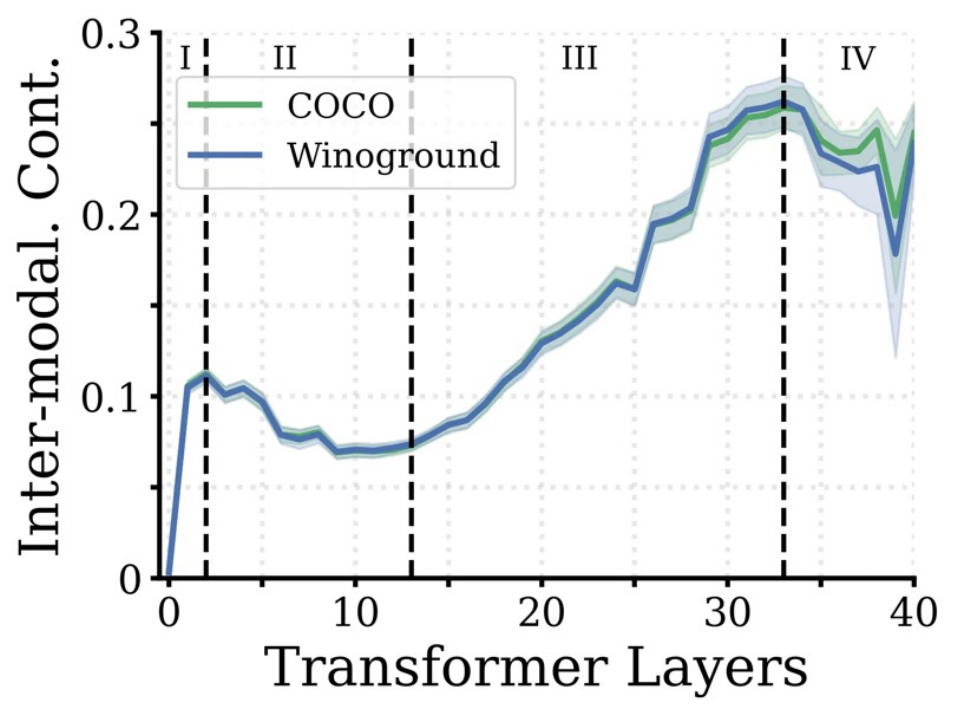}
    \label{fig:1a}
  \end{minipage}
  % \hfill % 添加适量空格
  \begin{minipage}[b]{0.48\linewidth}
    \centering
    \includegraphics[width=\textwidth]{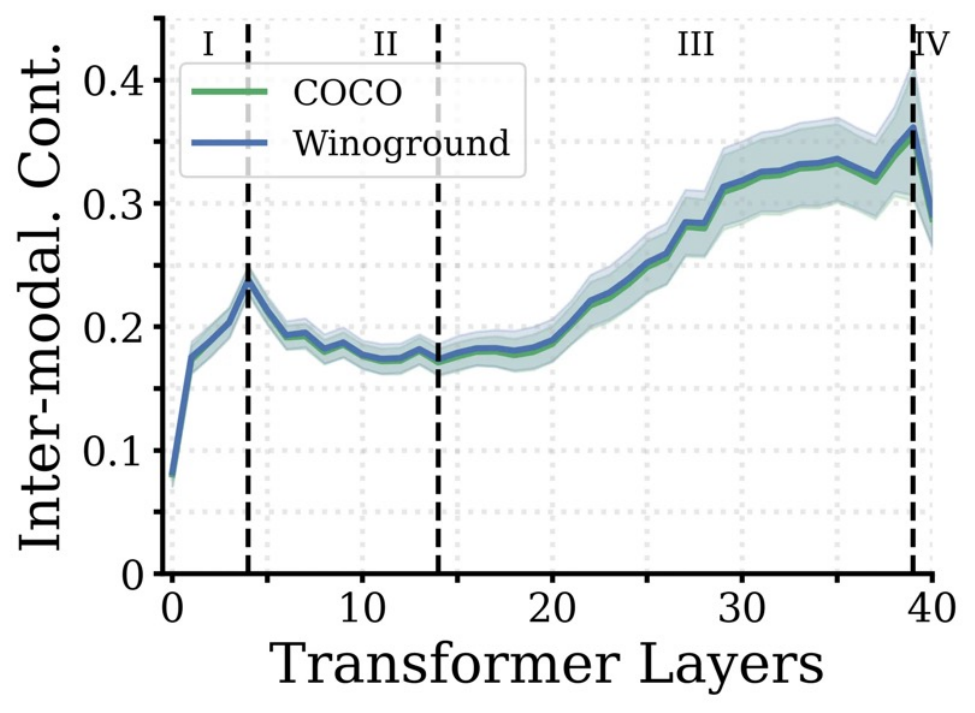}
    % \caption{Second Figure}
    \label{fig:1b}
  \end{minipage}
  \caption{Magnitude of inter-modal contextualization against layer depth. Four monotonical intervals indicate our proposed four-phase inference dynamics. A higher value indicates more aggressive multimodal interaction.\textbf{Left:} InstructBLIP. \textbf{Right:} LLaVA-v1.5.}
    \label{fig:1}
\end{figure}

% \begin{figure}
%     \centering
%     \includegraphics[width=0.8\linewidth]{Figures/Figure_1_Inter-modality_Contexutalization.pdf}
%     \caption{Magnitude of inter-modal contextualization against layer depth. Four monotonical intervals indicate our proposed four-phase inference dynamics. A higher value indicates more aggressive multimodal interaction.}
%     \label{fig:1}
% \end{figure}

\begin{figure*}
    \centering
    \includegraphics[width=1.0\linewidth]{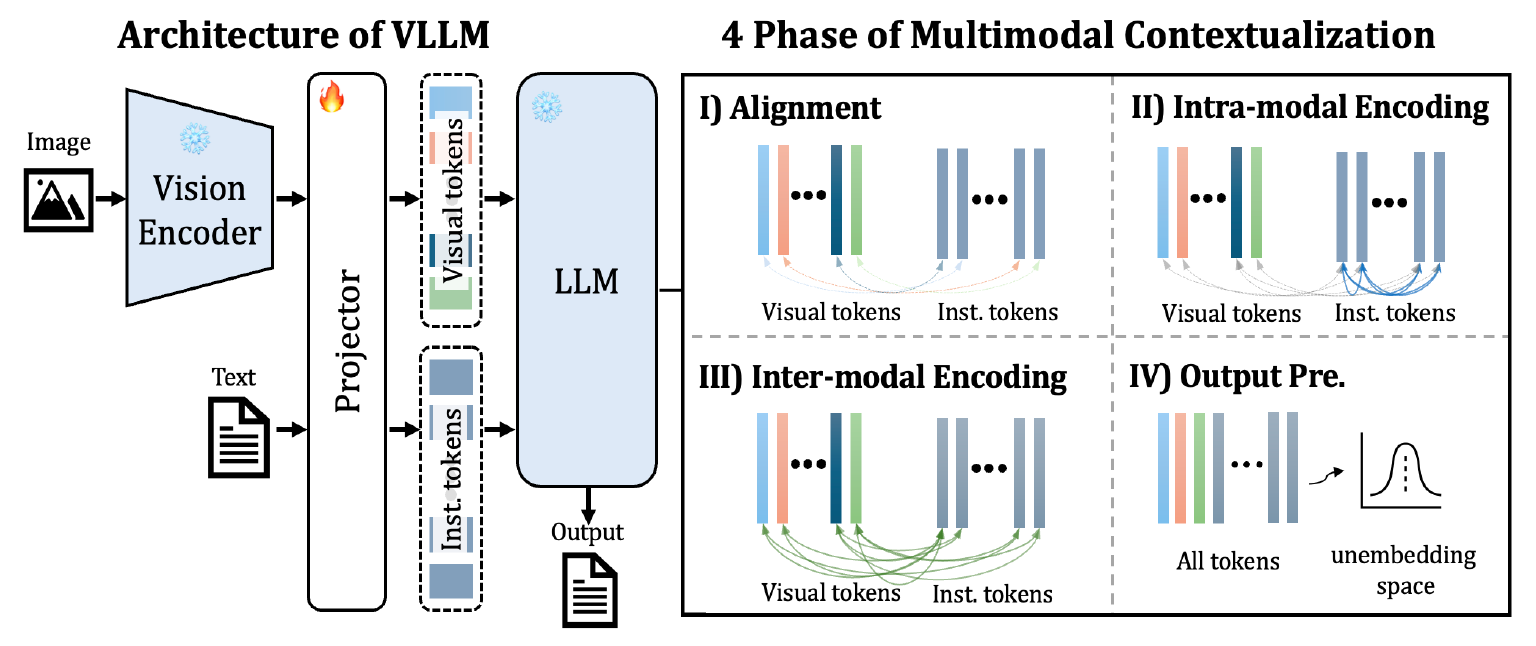}
    \caption{A four-phase diagram of feed-forward dynamics of LMs in VLLMs. (\textbf{I) Alignment} of two different feature spaces occurs. (\textbf{II) Intra-modal Encoding} is enhanced while cross-modal encoding is inhibited. (\textbf{III) Inter-modality Encoding} appears and strengthens. (\textbf{IV) Output  Preparation} requires hidden states to be aligned toward output embedding space.}
    \label{fig:2}
\end{figure*}

In this paper, we investigate the magnitude of contextualization~\citep{ethayarajh2019contextual} to characterize the multimodal interaction along LMs' layers.
Fig.~\ref{fig:2} illustrates the overview of our phase diagram of multimodal contextualization.
In detail, we look into the similarity among image tokens and instruction tokens, discovering that as inputs pass through successive layers of the Transformer, a phase diagram of multimodal contextualization with four monotonical intervals becomes apparent (Fig.~\ref{fig:1}).
\textbf{Phase I} shows an immediate ascent of contextuality, indicating an early alignment between two modalities. 
\textbf{Phase II} presents a slight decline in multimodal interaction meanwhile the active intra-modal encoding is found in Fig.~\ref{fig:3}, suggesting the concentration shift of model encoding.
\textbf{Phase III} shows a swift rise in similarity, indicating an incremental inter-modal interaction.
\textbf{Phase IV} presents the global reduction in inter-modal similarity, suggesting the model shifts focus away from multimodal interaction. Moreover, more fine-grained experiments are conducted to enhance our aforementioned findings of inference dynamics. Details are revealed in \S\ref{sec:attention} and \S~\ref{sec:logitlens}.

\section{Methodology} 
We investigate multimodal interaction by implementing the three following approaches, each delineating different aspects of the interaction process.
\subsection{Experiment Methods} % Method
\label{sec:method}

\noindent\textbf{Contextualization as Interaction Magnitude.} Inspired by~\citet{ethayarajh2019contextual}, we use cosine similarity as a measurement of contextuality to explore how hidden states from two different representation spaces interact in LMs.
In detail, for hidden states at each layer in LMs, we calculate the averaged cosine similarities among the line vectors as follows:
\begin{equation}
s^{(l)} = \frac{1}{mn} \sum_{i = 0}^m \sum_{j=0}^{n} \mathrm{cos}\left(v_i^{(l)}, w_j^{(l)}\right) \label{eq1},
\end{equation} 
\noindent where $m$ and $n$ indicate the number of visual tokens and text tokens. $v^{(l)}_{i}$ and $w^{(l)}_{j}$ refer to the visual token hidden states and text token hidden states at layer $l$ in LMs, respectively.

\noindent\textbf{Visualization via Norm-based Attention.}
% [Old attention is bad]
To investigate how the multimodal information interacts by multi-head attention mechanism, considering the faithfulness problem of attention score as an explanation \cite{clark2019does, serrano2019attention, jain2019attentionexplanation}, we use norm-based attention proposed by~\citet{kobayashi-etal-2020-attention}, which use the norm of multi-head attention's output transformation to scale the attention score, to measure the saliency assignment of Transformer. 

% We follow this to conduct our attention analysis on multimodal interaction.

\noindent\textbf{Verbalization via LogitLens.}
From another perspective, we wonder how visual tokens are converted into verbal concepts represented in language model space.
% how multimodal interaction affects visual tokens.
% whether the multimodal interaction facilitates the semantics of visual tokens converting to verbal concepts represented in language model space. 
Inspired by LogitLens~\cite{nostablog}, which interprets how LMs gradually conclude the final outputs via using the unembedding matrix to decode hidden states, in our paper, we use it to verbalize the operation by which visual tokens are aligned towards the language output space.
Specifically, by applying LM's output head to hidden states of visual tokens, we decode visual tokens into words at each layer of LMs in VLLM.

% Consider an arbitrary hidden state of visual token at layer $l$ of LMs in VLLM as $h_{vis}^{(l)}$, the vision version of LogitLens can be formulated as:

% \begin{equation}
%     \mathrm{LogitLens}\left(h_{vis}^{(l)}\right) = \mathrm{LayerNorm}\left[h_{vis}^{(l)}\right]W_{U},
% \end{equation}

% \noindent where $W_{U}$ is the unembedding matrix of LM. Note that the residual output of layer $l$ is ignored by setting the residuals to zero. 

\subsection{Experiment Settings}

\paragraph{Models.} We conduct experiments on two representative VLLMs. 1) InstructBLIP~\cite{10.5555/3666122.3668264}, which is extended from BLIP-2~\cite{li2023blip2bootstrappinglanguageimagepretraining}, introducing an instruction-aware Query Transformer to extract task-relevant image features tailored to the given instruction. 2) LLaVA-1.5~\cite{liu2024improvedbaselinesvisualinstruction} apply an MLP projection as the cross-modal connector on top of CLIP vision encoder, establishing new SOTA baselines across 11 VL benchmarks. A typical model architecture is shown in Fig.~\ref{fig:2} (Left).

\paragraph{Datasets.} For evaluation, we use COCO captions~\cite{chen2015microsoftcococaptionsdata} and Winoground datasets~\cite{thrush2022winogroundprobingvisionlanguage}. COCO is a commonly used image-caption dataset, contains 164K images, each annotated with five captions. Winoground is a carefully handcrafted probing dataset, comprising $400$ items, each including two pairs of images and corresponding captions.
%which challenges models to pair images with the minimum difference with corresponding captions correctly.

\paragraph{Other Details.} For each experiment, we randomly select $400$ images respectively from Winoground and COCO captions validation setas the testing samples.
We keep our prompt fixed across all experimental settings since our focus is observing the feed-forward dynamics of hidden states during inferences, rather than exploring the prompt engineering for VLLMs; therefore, we weaken the impact of the prompt variant.

\section{Phase Diagram of VLLM Inference Dynamics}
This section elaborates on our investigation of multimodal interaction along Transformer layers in VLLMs.
\S\ref{sec:multimodalCont.} reveals that multimodal interaction evolves as the Transformer layer goes deeper, introducing our main claim of the four-phase diagram of multimodal interaction during feed-forward calculation.
\S\ref{sec:attention} conducts the norm-based attention analysis, finding that intensive attention progressively emerges in the middle and later layers but significantly weakens in the last few layers, which further substantiates our findings.
Finally, \S\ref{sec:logitlens} uses LogitLens to verbalize how visual tokens are aligned towards LM output space across layers, presenting consistent trends with similarity-based results. 

\subsection{Four-phase Multimodal Contextualization}
\label{sec:multimodalCont.}
As described in equation (\ref{eq1}), we calculate the cosine similarity among hidden states from both modalities at each layer of LMs in VLLMs. 

The results are shown in Fig.~\ref{fig:1}, exhibiting a general upward trend as expected.
However, the curve does not monotonically increase throughout. In certain areas, such as layer 3 to 10, the curve briefly declines.
Therefore, we divide this similarity curve into four parts based on monotonicity and hypothesize that they correspond to four phases of the inference process (Fig.~\ref{fig:2} Right), i.e. (I) Alignment, (II) Intra-modal Encoding, (III) Inter-modal Encoding, and (IV) Output Preparation.

\begin{figure}
    \centering
    \begin{minipage}[b]{0.48\linewidth}
    \centering
    \includegraphics[width=\textwidth]{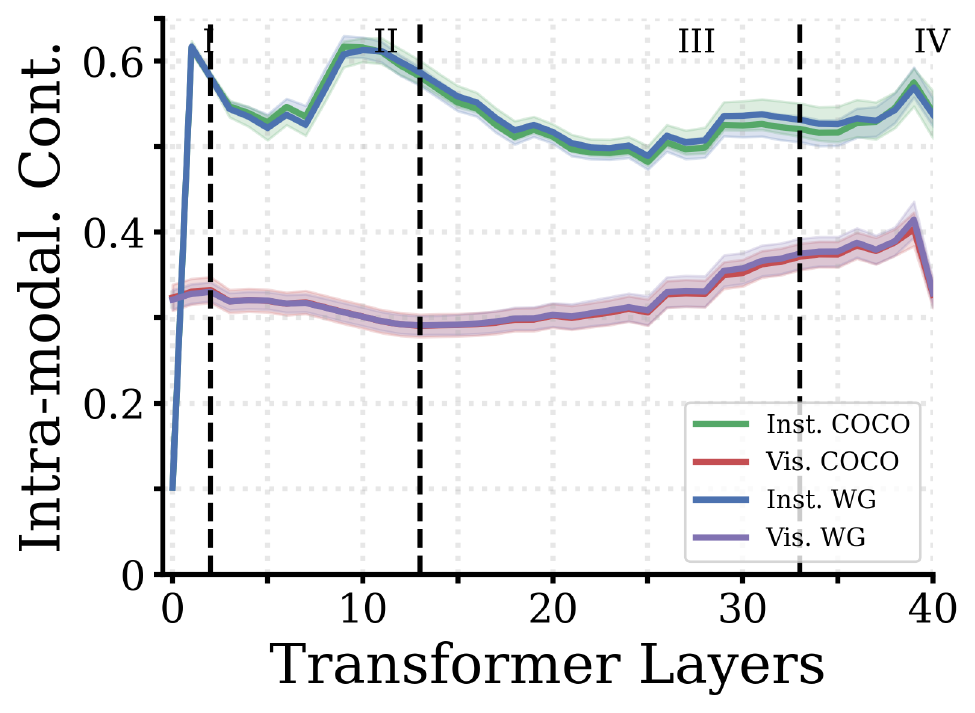}
    % \caption{First Figure}
    \label{fig:3a}
  \end{minipage}
  % \hfill % 添加适量空格
  \begin{minipage}[b]{0.48\linewidth}
    \centering
    \includegraphics[width=\textwidth]{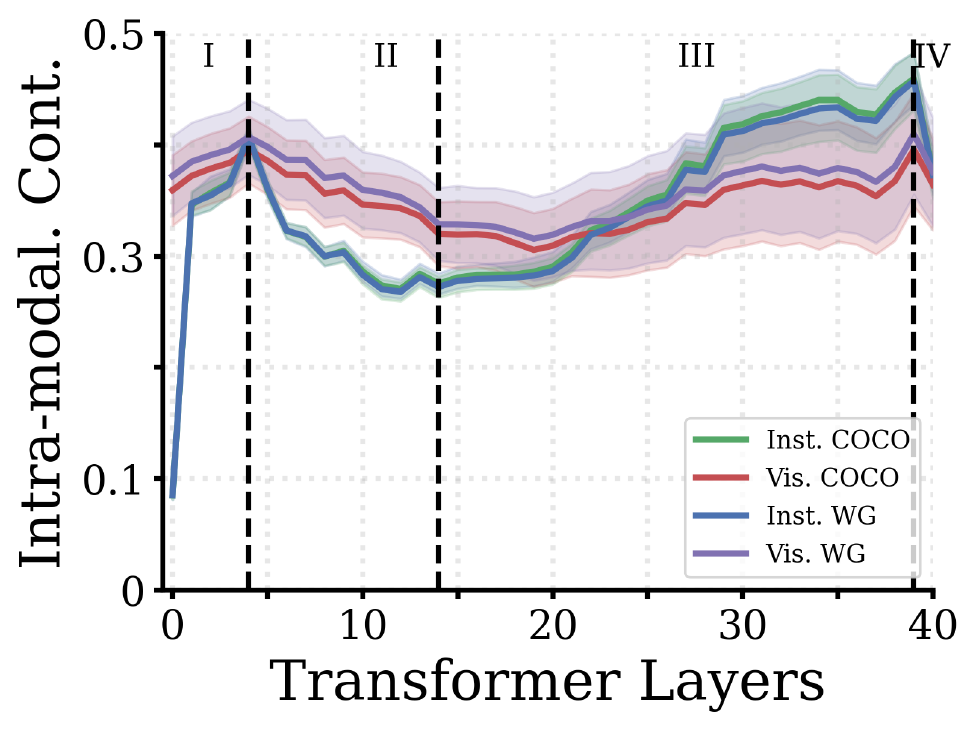}
    % \caption{Second Figure}
    \label{fig:3b}
  \end{minipage}
  \caption{Intra-modal contextualization of visual tokens and instruction tokens, respectively. Similarity values are averaged over randomly chosen $400$ images for each dataset.\textbf{Left:} InstructBLIP. \textbf{Right:} LLaVA-1.5}
    \label{fig:3}
\end{figure}

Besides, we conducted additional experiments by repeating the calculation of similarity for visual tokens and text tokens, separately.
As shown in Fig.~\ref{fig:3}, the contextuality within individual modality is much higher than that across modalities, indicating an active intra-modal encoding, which confirms our speculation.

\subsection{Visualization of Multimodal Interaction}
\label{sec:attention}

The findings of contextualization provide the overview of multimodal inference dynamics, but we further wonder how attention assignment is allocated during the feed-forward pass.
To answer this question, we visualize the norm-based attention assignment of the last text token.
\begin{figure*}
    \centering
    \begin{minipage}[b]{0.29\linewidth}
    \centering
    \includegraphics[width=\textwidth]{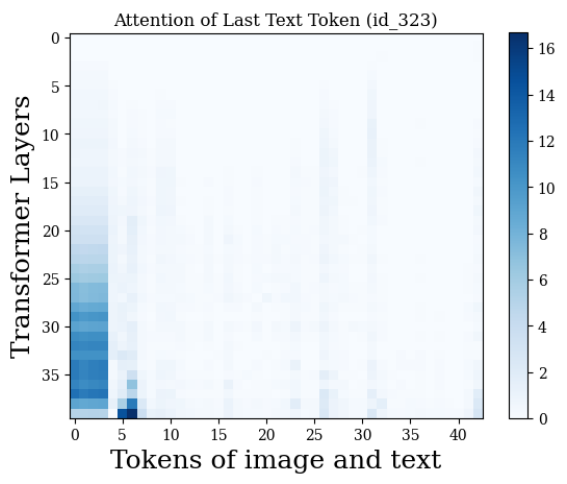}
    \label{fig:4a}
    \end{minipage}
    \begin{minipage}[b]{0.29\linewidth}
    \centering
    \includegraphics[width=\textwidth]{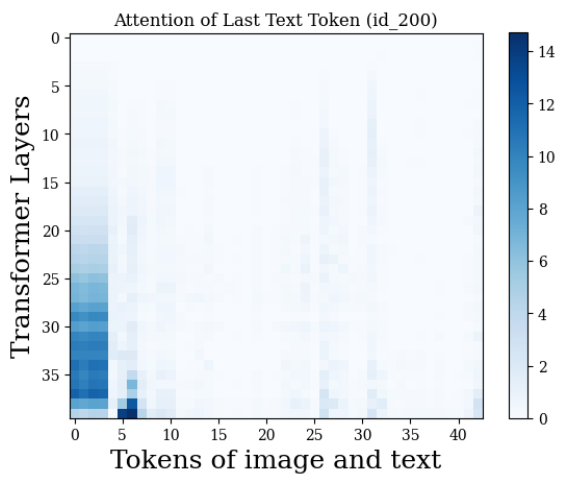}
    \label{fig:4b}
    \end{minipage}
    \begin{minipage}[b]{0.29\linewidth}
    \centering
    \includegraphics[width=\textwidth]{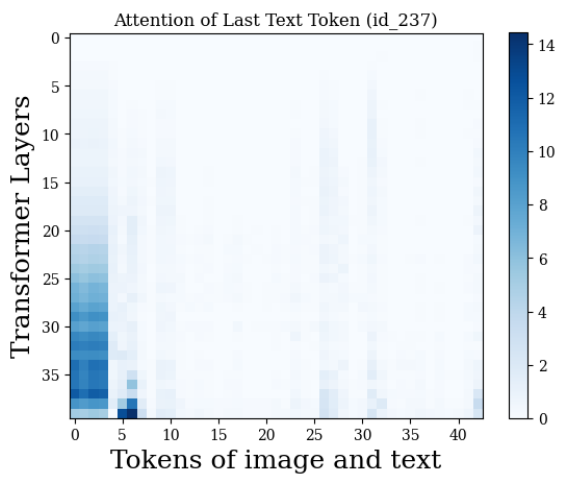}
    \label{fig:4c}
    \end{minipage}
    \begin{minipage}[b]{0.29\linewidth}
    \centering
    \includegraphics[width=\textwidth]{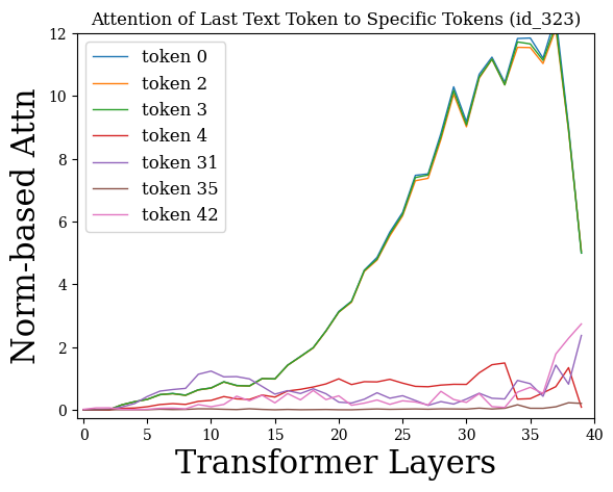}
    \label{fig:4d}
    \end{minipage}
    \begin{minipage}[b]{0.29\linewidth}
    \centering
    \includegraphics[width=\textwidth]{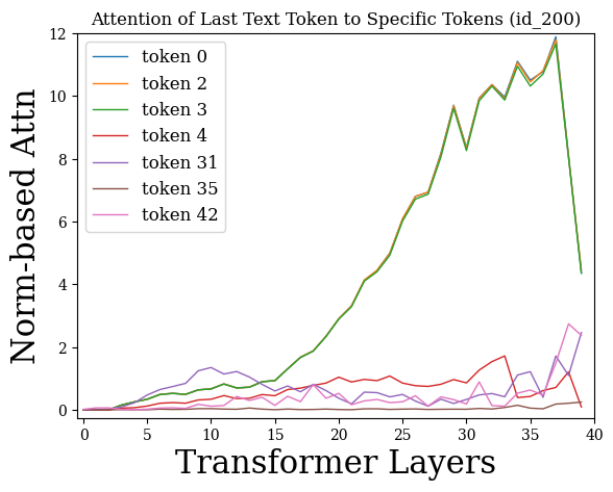}
    \label{fig:4e}
    \end{minipage}
    \begin{minipage}[b]{0.29\linewidth}
    \centering
    \includegraphics[width=\textwidth]{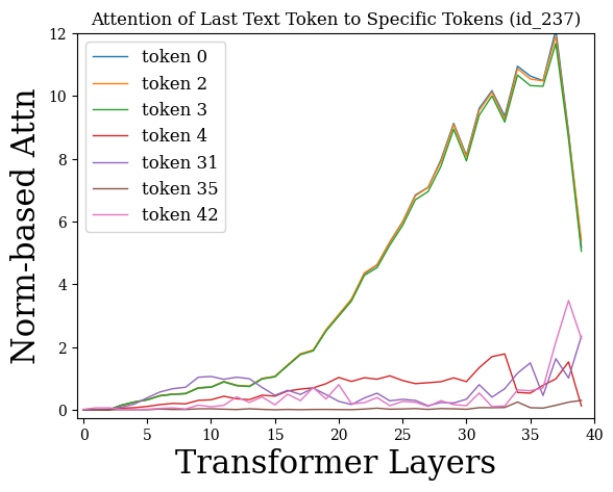}
    \label{fig:4f}
    \end{minipage}

    \caption{Visualization of norm-based attention analysis (id\_$323$ refers to a randome chosen image). \textbf{Above:} Heatmaps showcase norm-based attention of the last text token to other tokens across layers. The more the darkness, the more it draws attention. \textbf{Below:} Norm-based attention of last text token to specific tokens over layers.}
    \label{fig:4}
\end{figure*}

As shown in Fig.~\ref{fig:4}, the saliency assignments of the last text token continuously accumulate in the middle and deeper layers but suddenly drop in the last few layers, showing a partially similar changing pattern as found in \S\ref{sec:multimodalCont.}. Note that we only showcase the norm-based attention results of two images (id\_$323$, id\_$200$), because we found almost identical pattern holds for all heatmaps of $50$ randomly selected images.

In addition, we found that the first several tokens received far more attention than other tokens. In particular, starting from the middle layers, their attention scores become substantially stronger than those of other tokens as the layer deepens. To explain this observation more intuitively, we plot the norm-based attention scores of the last text token to several chosen non-trivial tokens against the layer number. As shown in Fig.~\ref{fig:5} Below, we found that across all layers, the first five tokens are allocated significantly more attention than the others.

\subsection{Verbalization of Visual Tokens}
\label{sec:logitlens}
We use LogitLens~\citep{nostablog} to verbalize the procedure by which visual tokens are converted into word tokens in LM space.

\begin{figure}
    \centering
    \includegraphics[width=0.8\linewidth]{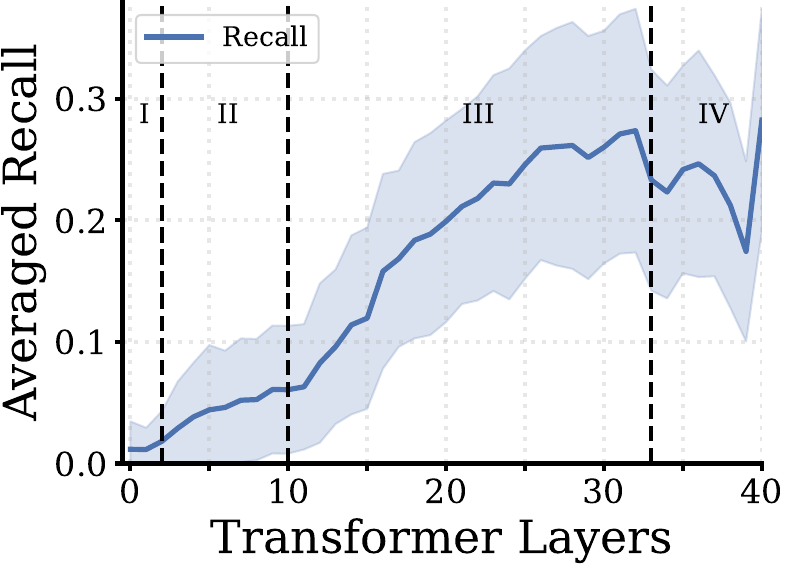}
    \caption{Averaged recall of decoded LogitLens words. Results are averaged over 2 datasets, each of which consists of randomly chosen 400 images.}
    \label{fig:5}
\end{figure}

Fig.~\ref{fig:5} shows the averaged recall score of the decoded LogitLens words using annotated captions as ground truth.
We find that the recall coheres well with the phase diagram of multimodal contextualization, especially Phase III and IV.

Specifically, we observe the following pattern across both datasets. In the middle layers ($10$-$32$), the recall gradually increases (from $0.1$ to $0.39$), implying that the hidden states of visual tokens are continuously evolving, and producing more caption-related words.
Then, from the layer around $33$, the recall starts decreasing.
We therefore speculate that the LM acts as an encoder in the middle layers, focusing on encoding image semantics, and then afterward the LM tends to more focus on preparing the output, rather than encoding.
This, from another perspective, complementarily confirms our four-phase of multimodal interaction during reasoning dynamics.

% [Figure Major revision: 1. Average or 2. Stat. Measure revision]

\section{Conclusion}
This paper proposes to utilize contextualization as a measurement to explain multimodal interaction in LMs of VLLMs.
By incorporating other investigation methods, our extensive experiments reveal a four-phase diagram of inference dynamics in terms of multimodal interaction during the model's feed-forward pass.
In detail, \textbf{(I)} Early alignment of hidden states from two different spaces appears in the first few layers; \textbf{(II)} Instruction encoding is observed in the middle layer; \textbf{(III)} Inter-modal interaction is progressively enhanced in deeper layers; \textbf{(IV)} Hidden states are aligned towards the output space of LM in last few layers. Our findings contribute to uncovering the inner workings of VLLMs.

% \paragraph{Anisotropy in multimodal contextualization.} Anisotropy is a phenomenon of word contextualization where the representation of words occupies a narrow cone in the embedding space instead of being uniformly distributed. Our results on multimodal interaction experiments showed that more aggressive contextualization occurs in deeper layers, indicating higher anisotropic as the layer goes deeper.

% try others like Wu-Palmer similarity (Wup)~\cite{wu1994verb}, which calculates the distance between the ground truth word and the decoded logit-lens words in the WordNet taxonomy, offering a way to measure ‘how close’ a word was to the correct answer.

\section*{Limitations and Future work}
% \paragraph{Fewer Prompt Variant.} To focus on investigating the ability of VLLMs to understand image content, we fix our prompt in all three experiments.
\textbf{More VLLMs Variants.} Although our experiments examine different aspects of multimodal interaction, testing these properties using one model would not be enough. Other VLLMs with the similar architecture (Fig.~\ref{fig:2}), e.g. LLaVA~\cite{liu2023visualinstructiontuning, liu2024improvedbaselinesvisualinstruction}, MiniGPT-4~\cite{zhu2023minigpt4enhancingvisionlanguageunderstanding}, should be included in the testing scope.

\noindent\textbf{Ablation on Certain Non-trivial Visual Tokens} Our norm-based attention analysis (Fig.~\ref{fig:4}) demonstrates that the first $5$ visual tokens are assigned far more attention compared with other visual tokens across all layers, which displays an interesting property for visual tokens in LMs during VLLMs' inference. Future work could use this property to do a broader exploration of parameter reduction during inferences. For example, feeding only the first $5$ visual tokens into VLLMs to observe the performance on VL tasks.

\noindent\textbf{Early-exit Decoding.} Given the widespread belief that enhanced contextuality between visual and textual modality benefits VL tasks, a potential use case of our findings could be early-exit decoding, which leverages hidden states of late layers for decoding rather than that of the final layer.

% Besides, from the perspective of complementing this work for revealing the mechanism of multimodal interaction, future work could analyze mutual information between representations.
% \section*{Acknowledgments}

\bibliography{custom}

% \appendix

% \section{Example Appendix}
% \label{sec:appendix}

% This is an appendix.

\end{document}